\documentclass{article}
\usepackage{spconf,amsmath}
\usepackage{epsfig}
\usepackage{epstopdf}
\usepackage{subfig}
\usepackage{graphicx}
\usepackage{float}
\usepackage{algorithm,algorithmic}
\usepackage{url}
\usepackage{color}
\usepackage{setspace}

\title{Deep Networks with Shape Priors for Nucleus Detection\vspace{-5pt}}
\name{\vspace{-3pt}\hspace{-4mm}Mohammad Tofighi, Tiantong Guo, Jairam K.P. Vanamala$^{\dagger}$, and Vishal Monga\vspace{-1pt}}
	
\address{\hspace{-2mm}Department of Electrical Engineering, Pennsylvania State University, University Park, PA \\ 
	\hspace{-3mm}$^{\dagger}$Center for Molecular Immunology and Infectious Disease, Pennsylvania State University, Univ. Park, PA \\ 
	\hspace{-2mm}Emails: tofighi@psu.edu, tiantong@psu.edu, juv4@psu.edu$^{\dagger}$, vmonga@engr.psu.edu\vspace{-5pt}}

\begin{document}
%
\maketitle
\begin{abstract}
	\pretolerance=5000
	\tolerance=8000
	\emergencystretch=10pt
Detection of cell nuclei in microscopic images is a challenging research topic, because of limitations in cellular image quality and diversity of nuclear morphology, i.e. varying nuclei shapes, sizes, and overlaps between multiple cell nuclei. This has been a topic of enduring interest with promising recent success shown by deep learning methods. These methods train for example convolutional neural networks (CNNs) with a training set of input images and known, labeled nuclei locations. Many of these methods are supplemented by spatial or morphological processing. We develop a new approach that we call Shape Priors with Convolutional Neural Networks (SP-CNN) to perform significantly enhanced nuclei detection. A set of canonical shapes is prepared with the help of a domain expert. Subsequently, we present a new network structure that can incorporate `expected behavior' of nucleus shapes via two components: {\em learnable} layers that perform the nucleus detection and a {\em fixed} processing part that guides the learning with prior information. Analytically, we formulate a new regularization term that is targeted at penalizing false positives while simultaneously encouraging detection inside cell nucleus boundary. Experimental results on a challenging dataset reveal that SP-CNN is competitive with or outperforms several state-of-the-art methods.\vspace{-5pt}
\end{abstract}
\begin{keywords}
	Cellular imagery, nucleus detection, deep learning, convolutional neural networks, shape priors\vspace{-5pt}
\end{keywords}
\section{Introduction}\vspace{-5pt}
\label{sec:intro}

Automatic analysis of cellular imagery to determine nuclei locations is a centrally important problem in diagnosis of several medical conditions including tumor and cancer detection \cite{SC_CNN}. Traditionally, a plethora of feature extraction and morphological processing techniques were developed \cite{veta2014breast}. Recently, there has been a surge of deep learning methods that have shown unprecedented accuracy in nuclei detection. 
\begin{figure*}[ht!]
	\centering
	\includegraphics[width=0.88\linewidth]{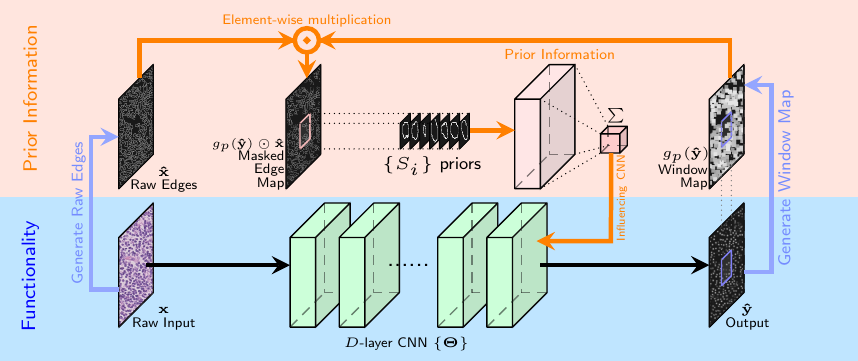}\vspace{-1pt}
	\caption{\ninept SP-CNN illustration. There are two parts of the SP-CNN: Functionality part (blue) and Prior Information part (orange). The Functionality part consists of one $D$-layer CNN that takes input raw image $\mathbf{x}$ and generates the detected labels for cell nuclei $\mathbf{\hat{y}}$. The prior information part computes the prior cost term as in Eq. (\ref{eq:sp}) and feeds the information into the (learning of the) CNN to guide it towards enhanced nuclei detection. Note the prior information part generates a regularization term used in training the $D$ learnable layers of the CNN and {\em only} the functionality part of the network is applied to a test image.}\vspace{-8pt}
	\label{fig:diagram}
\end{figure*}

\textbf{Related work:} Some of the earliest attempts at cell nuclei detection involved tailored feature extraction and morphological processing \cite{LIPSyM,veta2014breast,al2010improved,Ali_SP_2012}. One key limitation of these approaches is that the best features for nuclei detection are rarely readily apparent. Further, often the designed techniques are too specific to choice of dataset and not versatile. Because of their ability to perform feature discovery and generalizable inference, deep learning methods have recently become popular for this problem. For instance, Cruz-Roa \textit{et al.} \cite{Cruz-Roa2013} showed that a deep learning architecture for nuclei detection outperforms methods based on different image representation strategies e.g. bag of features, canonical and wavelet transforms. Xie \textit{et al.} \cite{Xie2015}, proposed a structural regression model for CNN, where a cell nuclei center is detected if it has the maximum value in the proximity map. In \cite{Xu2016}, Xu \textit{et al.} proposed a cell detection method based on a stacked sparse autoencoder, where it learns high level features of cell centroids and then a softmax classifier is used to separate the nuclear and non-nuclear image patches. Sirinukunwattana \textit{et al.} proposed SC-CNN \cite{SC_CNN} which uses a regression approach to find the likelihood of a pixel being the center of a nucleus. In SC-CNN, the probability values are topologically constrained in a way that in vicinity of nuclei center the probability is higher. Another recent approach \cite{Xing2016} uses a combination of well-known traditional CNNs for cell nuclei segmentation and dictionary learning techniques for refining results.

In this paper, we take a different approach by exploiting a prior understanding of the shape of the nuclei, which can be obtained in consultation with a medical domain expert. We note that
shapes have played an important role in medical image segmentation. Leventon \textit{et al.} \cite{Leventon_SP_2000} proposed a method for medical image segmentation by incorporating shape information into the geodesic active contour method. Ali \textit{et al.} \cite{Ali_SP_2012} incorporated prior shape information into boundary and region based active contours for accurate segmentation of cells. 



\noindent \textbf{Motivation:}
While existing deep learning approaches for cell nuclei detection are promising, our goal is to fundamentally alter the learning of the network by enriching it with domain knowledge provided by a medical expert. In particular, we propose to incorporate expert designed nuclei Shape Priors with Convolutional Neural Networks (SP-CNN). 

Our \textbf{contributions} include: 1) a novel network structure that can incorporate `expected behavior' of nucluei shapes via two components: {\em learnable} layers that perform the nucleus detection and a {\em fixed} processing part that guides the learning with prior information. Three sources contribute to generating the prior: network output, raw edge map from the input image, and a set of predefined shapes prepared by the medical expert, 2) analytically, we formulate a new regularization term that is targeted at penalizing false positives while simultaneously encouraging detection inside cell boundary. We carefully design this term so it is differentiable w.r.t. the output and hence the network parameters, enabling tractable learning through standard back-propagation schemes. \vspace{-3pt}

\section{Shape Priors with Convolutional Neural Networks (SP-CNN)}\vspace{-3pt}

\subsection{CNN for Nucleus Detection}\vspace{-3pt}
SP-CNN detects the cell nucleus using a regression CNN. In regression networks, the goal is to obtain a function that interprets the relationship between the input image $\mathbf{x}$ and ground truth labeled image $\mathbf{y}$. The network is modeled by parameters set $\mathbf{\Theta}=\{\mathbf{W}, \mathbf{b}\}$, where $\mathbf{W}$ and $\mathbf{b}$ are denote respectively the weights and bias of all layers combined. The CNN is learned by solving the well-known optimization problem \cite{SC_CNN,RegFuncRef1,RegFuncRef2}:\vspace{-2mm}
\begin{equation}
\mathbf{\Theta} = \arg\min\limits_{\mathbf{\Theta}}\|f(\mathbf{x}; \mathbf{\Theta}) - \mathbf{y}\|_2^2
\label{eq:CNN_func}\vspace{-2mm}
\end{equation}
where $f(\mathbf{x;\Theta})$ represents the non-linear mapping of the CNN that generates the detection maps $\mathbf{\hat{y}}$ \footnote{\noindent We work with soft labels $\mathbf{y}$ and $\mathbf{\hat{y}}$, i.e.\ $\mathbf{y}, \mathbf{\hat{y}}$ can take values in the range $[0,1]$. In practice, $\mathbf{y}$ is obtained by processing the binary image (0 or 1 at each pixel) of ground truth nuclei locations as in \cite{SC_CNN}; also see Section \ref{sec:experiments}.}. Each of the $D$ CNN layers comprises of a convolutional layer followed by an activation function, which is a Rectified Linear Unit (ReLU) \cite{relu}.

\begin{figure}
	\begin{center}
	\vspace{-4pt}
		\includegraphics[width=85mm, trim={0 0 0 0},clip]{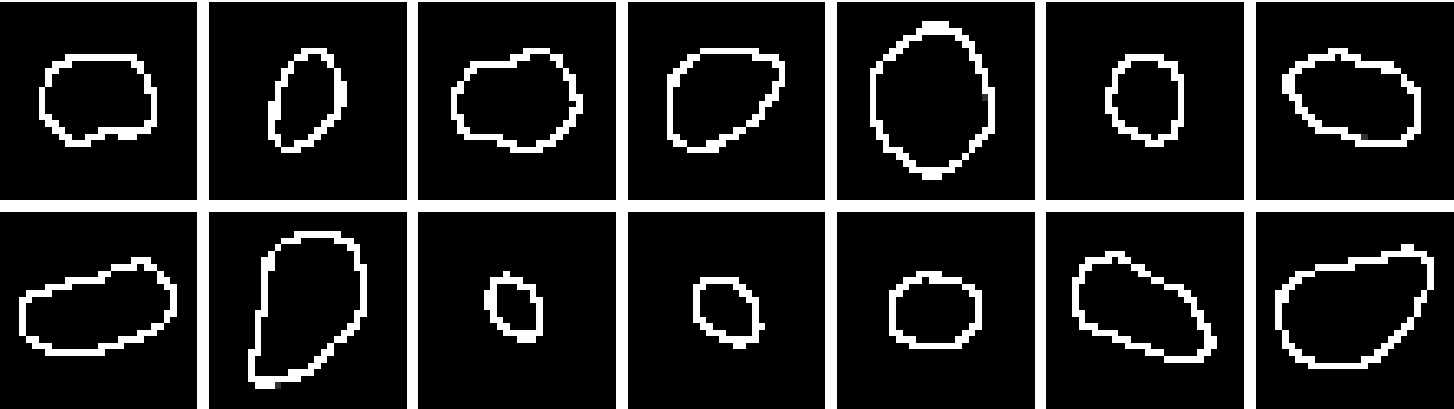}\vspace{-5pt}
		\caption{\ninept Samples of handcrafted cell nuclei shapes from colon tissues}
		\label{fig:shapes}
	\end{center}\vspace{-25pt}
\end{figure}

\subsection{Deep Networks With Shape Priors}\vspace{-2mm}
As discussed in Section \ref{sec:intro}, we now incorporate the prior information about the shape of the cell nuclei into the training of the CNN. Ideally, the labels produced by the CNN should lie inside of the nuclei boundaries. We set up a regularization term to explicitly encourage the learned network which encourages detection inside the nucleus boundary while simultaneously penalizing false positives. The regularizer is based on a set of shape priors developed with the help of domain expert and given by:
$\mathbf{S} = \{\mathcal{S}_i | i = 1, 2, \dots, n\}$.

\begin{figure}
	\centering
	\includegraphics[width=0.49\textwidth]{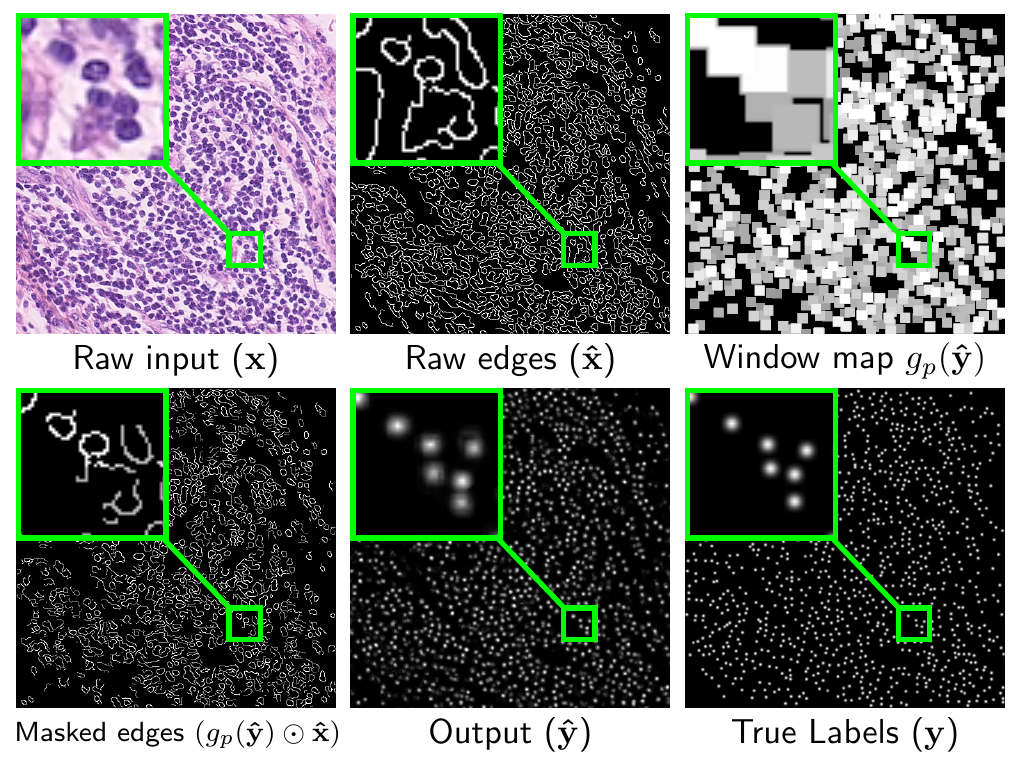}
	\vspace{-18pt}
	\caption{\ninept Images in each step of SP-CNN.}\vspace{-10pt}	\label{fig:sample_patches}
\end{figure}

For each dataset, multiple training images are analyzed by a medical expert to hand label the nuclei boundaries. A set of $n$ representative shapes is then hand selected by the medical expert to form the set $\mathbf S$. Some examples of the nucleus shape priors are shown in Fig. \ref{fig:shapes}. These are corresponding to colon tissue images -- detailed explanation is provided in Section \ref{sec:experiments}.  To construct a meaningful regularization term emphasizing shape priors, we need the nucleus boundary information of the input raw image $\mathbf x$. We employ the widely used Canny edge detection filter \cite{Canny} to generate the raw edge image $\mathbf{\hat{x}}$ with edges labeled as $1$ and background as $0$, as shown in Fig. \ref{fig:sample_patches}. Note that the raw edge image $\mathbf{\hat{x}}$ is only used during the training process.

We now define the regularization term that captures shape priors:
\begin{equation}
\sum\limits_{i=1}^{n} \|(g_{p}(\mathbf{\hat{y}})\odot \mathbf{\hat{x}}) \ast \mathcal{S}_i \|_2^2\label{eq:sp}
\end{equation}
where the $g_{p}(\cdot)$ denotes the max pooling operation on $\mathbf{\hat{y}}$ with window size $p$. Based on Eq. (\ref{eq:sp}), the computation of the shape priors cost term consists of three steps as shown in Fig. \ref{fig:diagram}'s prior information (orange) part: 1) the CNN output $\mathbf{\hat{y}}$ is first thresholded by $T_p = 0.2$ to eliminate the background noise and then max pooled by $g_p(\cdot)$ with stride of 1 and the `SAME' padding scheme. This results in a window map $g_p(\mathbf{\hat{y}})$ that has $p\times p$ window centered at each location within the soft detected region. 
As expected, a window with higher numerical value will result if the detected label values (in $\mathbf{\hat{y}}$) are correspondingly higher (closer to 1), 2) the window map $g_p(\mathbf{\hat{y}})$ is then multiplied with the raw edge image $\mathbf{\hat{x}}$ element-wise. This step serves to mask out the edges from $\mathbf{\hat{x}}$ that surround the detected location in $\mathbf{\hat{y}}$, as shown in Fig. \ref{fig:sample_patches}, 3) the masked edge image $(g_{p}(\mathbf{\hat{y}})\odot \mathbf{\hat{x}})$ is convolved with the shape priors in set $\mathbf{S}$ to generate a measurement of how well does the detection fit inside the nucleus shape. If  $\mathbf{\hat{y}}$ has more labels predicted inside the nucleus boundary, Eq. (\ref{eq:sp}) will produce a higher value.

Note that the effect of the shape prior is captured by a negative regularization term since the goal is to maximize (and not minimize) correlation with `expected shapes'. Overall, the cost function of the SP-CNN is given by:\vspace{-7pt}
\begin{equation}
\hspace{-1pt}\mathbf{\Theta} = \arg\min\limits_{\mathbf{\Theta}}\|f(\mathbf{x}; \mathbf{\Theta}) - \mathbf{y}\|_2^2 - \lambda\sum\limits_{i=1}^{n}  \|(g_{p}(\mathbf{\hat{y}})\odot \mathbf{\hat{x}}) \ast \mathcal{S}_i \|_2^2
\label{eq:costFunc}\vspace{-7pt}
\end{equation}
where $\lambda$ is the trade-off parameters between the detection fidelity term and the value representing the effect of the shape prior. Note that $\mathbf{\hat{y}}:=f(\mathbf{x}; \mathbf{\Theta})$, thus the shape prior cost term is effected by the network parameters and also introduces gradient terms that updates the network parameters during the training process using back-propagation \cite{lecun2015deep}. Please refer to our supporting document for detailed derivations \cite{webpage}.


\vspace{-5pt}
\section{Experimental Results}\vspace{-1mm}
\label{sec:experiments}
\subsection{Data Preparation}\vspace{-1pt}
\label{sec:data}
We train and test SP-CNN on publicly available dataset of \cite{SC_CNN} which includes $100$ H\&E stained histology images of colorectal adenocarcinomas. There are a total number of $29756$ nuclei marked at the nucleus center (please refer to Sec. VII.A. of \cite{SC_CNN} for more information).

We construct $\mathbf{y} \in [0,1]$ by processing the binary image of ground-truth nuclei center locations, which has 1 at the nucleus center and 0 elsewhere. This is accomplished by convolving the said ground-truth binary image with a zero mean Gaussian ($\sigma = 2$) filter of size $7 \times 7$. Then the (luminance) input image, the raw edge image, and the labeled image form a training tuple $(\mathbf{x}, \mathbf{\hat{x}}, \mathbf{y})$; patches of size $40\times 40$ are extracted and used for training. There are void regions in the raw image $\mathbf{\hat{x}}$, which do not include nuclei, as a result the label image $\mathbf{y}$ is also void. These redundant void regions can mislead the network while wasting computation. To avoid this, we introduce a procedure to eliminate the training patches which are empty in $(\mathbf{x}, \mathbf{\hat{x}}, \mathbf{y})$. 

\subsection{SP-CNN Parameters}
\label{sec:params}
The SP-CNN uses a CNN with $D=6$ layers\footnote{We chose 6 layers to be consistent with competing deep networks which also employ 6-8 layers \cite{SC_CNN,Xing2016}.} with `SAME' padding scheme; its configuration details are provided in Table. \ref{tab:network}. We use $n = 64$ different nuclei shapes in our shape set $\mathbf{S}$. Each shape is described by a $20 \times 20$ patch for the dataset in \cite{SC_CNN}. The active part of the shapes are labeled as 1 and 0 otherwise. All the parameters used in SP-CNN are chosen by cross validation \cite{monga2017,haddadpour2012coordination}. Most important of them are: trade-off value $\lambda~=~5e-7$, pooling window size $p = 11\times 11$, weight decay parameter = $1e-5$, learning rate decay = $0.75$.
\begin{table}[t]
	\centering
	\caption{\ninept The Configuration of The CNN Used in SP-CNN}\vspace{-3pt}
	\label{tab:network}
\resizebox{.98\linewidth}{!}{
		\begin{tabular}{cccc}
			\hline\hline
			Layer No. & Layer Type & Filter Dimensions & Filter Numbers\\ \hline
			1 & Conv. + ReLU & $5\times 5\times 1$ &$64$ \\ 
			2 & Conv. + ReLU & $3\times 3\times 64$ &$64$ \\ 
			3 & Conv. + ReLU & $3\times 3\times 64$ &$64$ \\ 
			4 & Conv. + ReLU & $3\times 3\times 64$ &$64$ \\ 
			5 & Conv. + ReLU & $3\times 3\times 64$  &$64$\\ 
			6 & Convolutional & $3\times 3\times 64$  &$1$\\ \hline\hline
		\end{tabular}}\vspace{-10pt}
\end{table}

\subsection{Assessment Methods And Comparisons}\vspace{-1pt}
\label{sec:assessment}
Note the output of SP-CNN $\mathbf{\hat{y}} \in [0,1]$, which is first processed via a thresholding operation with a pre-determined threshold $T$. Local maxima of the resulting thresholded image are identified as detected nuclei locations.\vspace{-0pt}

To evaluate the detected locations against the true ones, we need some tolerance since it is unlikely that they will exactly match.
This is handled in the literature by defining a golden standard region as a region of 6 pixels around each ground-truth nuclei center as in \cite{SC_CNN} for its dataset. A detected nuclei location is considered to be true positive ($TP$), if it lies inside this region, otherwise it is considered to be false positive ($FP$), and the ones that are not matched by any of golden standard regions are considered to be false negative ($FN$). For quantitative assessment of SP-CNN and comparison with other methods we use Precision ($P$), Recall ($R$), and F1 score ($F1$), which are defined as:
$\small P = \frac{TP}{TP + FP}, ~R = \frac{TP}{TP + FN}, ~\text{and}~ F1 = \frac{2PR}{P+R}.$

\begin{figure}[t]
	\centering
	\includegraphics[width=0.48\textwidth]{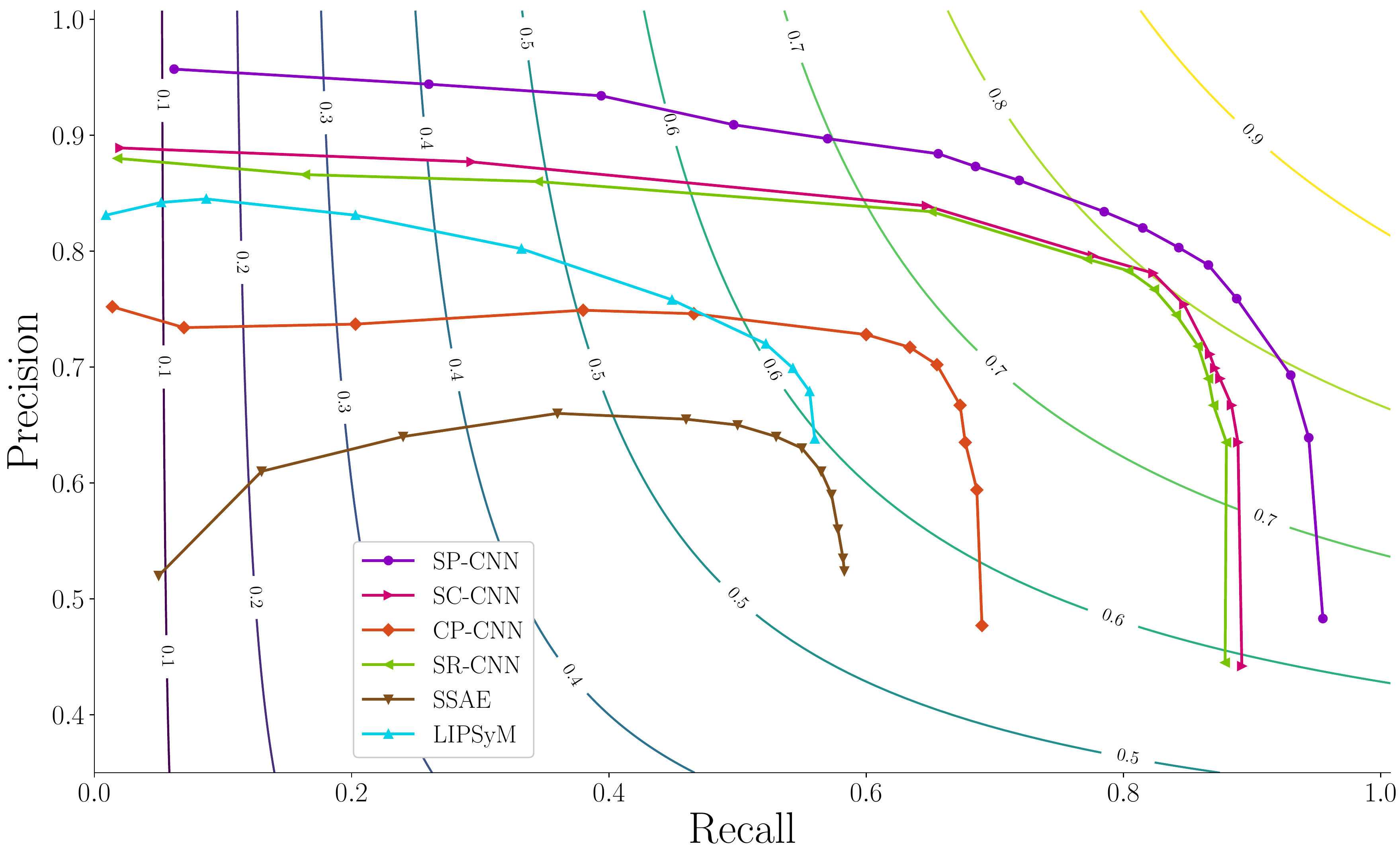}\vspace{-1mm}
	\caption{\ninept Precision-recall curve.}
	\label{fig:PRcurve}\vspace{-5pt}
\end{figure}

As is common, $T$ is varied to generate a Precision-Recall curve. The Precision-Recall curve for averaged values over all images in the test set from the dataset in \cite{SC_CNN} is plotted in Fig. \ref{fig:PRcurve}.
For consistency, results for the proposed SP-CNN are based on using the same assessment procedure as in \cite{SC_CNN}, which used a $50\text{-}50$ split of training vs. test images (using the official assessment source codes of \cite{SC_CNN} provided by the paper's author). Fig. \ref{fig:PRcurve} essentially compares SP-CNN against state of the art deep learning methods: SC-CNN \cite{SC_CNN}, CP-CNN\cite{SC_CNN}, SR-CNN\cite{Xie2015}, SSAE\cite{Xu2016}, and two other popular feature and morphology based methods LIPSyM\cite{LIPSyM}, and CRImage\cite{CRImage}.
For fairness of comparison, $P, R,$ and $F1$ results for the aforementioned competing methods are obtained directly from comparisons already reported in \cite{SC_CNN}. 
 Fig. \ref{fig:PRcurve} reveals that SP-CNN achieves the best Precision-Recall curve. 

To obtain a single representative figure of merit for each method, we chose the threshold value which maximizes the F1-score correspondingly for each method. These `best F1-scores' for each method are then reported in Table \ref{tab:results}. We also include the results of SP-CNN without shape priors in Table \ref{tab:results} to show the value of the proposed regularizer in Eq. (\ref{eq:costFunc}). 

\begin{table}[b!]
	\centering
	\caption{\ninept Nucleus detection results for dataset \cite{SC_CNN}}\vspace{-1pt}
	\label{tab:results}
	\resizebox{0.98\linewidth}{!}{
		\begin{tabular}{cccc}
			\hline	\hline
			 Dataset from \cite{SC_CNN} & Precision & Recall & F1 score \\ \hline
			SP-CNN & \textbf{0.803} & \textbf{0.843} & \textbf{0.823} \\ 
			SP-CNN ({\it without shape priors})& 0.757 & 0.818 & 0.786 \\ 
			SC-CNN \cite{SC_CNN} & 0.781 & 0.823 & 0.802 \\
			CP-CNN\cite{SC_CNN} & 0.697 & 0.687 & 0.692 \\ 
			SR-CNN\cite{Xie2015} & 0.783 & 0.804 & 0.793 \\
			SSAE\cite{Xu2016} & 0.617 & 0.644 & 0.630 \\
			LIPSyM\cite{LIPSyM} & 0.725 & 0.517 & 0.604 \\ 
			CRImage\cite{CRImage} & 0.657 & 0.461 & 0.542\\ \hline\hline
		\end{tabular}}
	\end{table}

	\begin{figure}[ht!]
		\centering
		\includegraphics[width=0.5\textwidth]{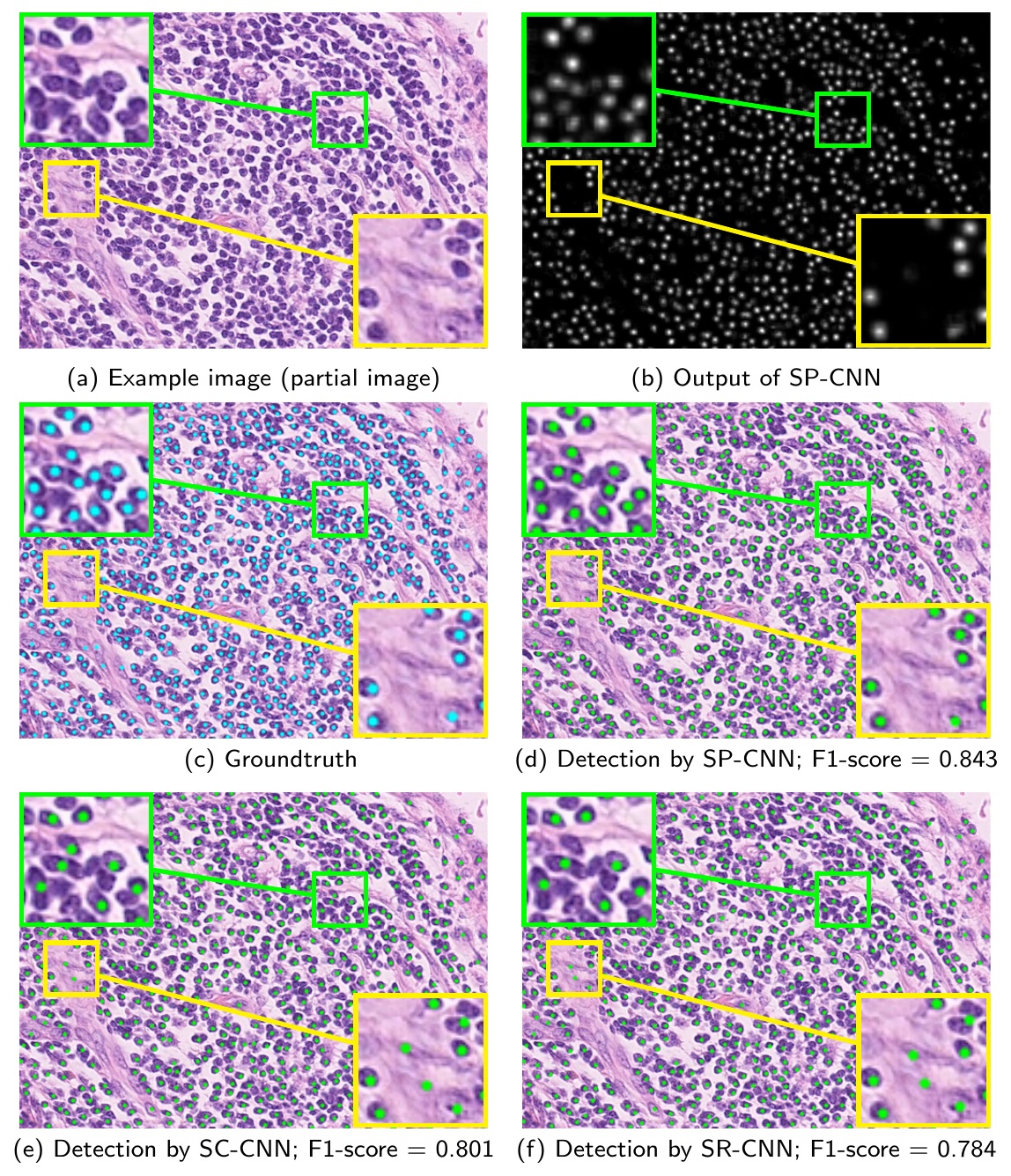}\vspace{-3mm}
		\caption{\ninept Example detection result for SP-CNN on \cite{SC_CNN}'s dataset.}
		\label{fig:example_result}\vspace{-5mm}
	\end{figure}

A visual illustration of nuclei detection results is presented in Fig. \ref{fig:example_result} for an example test image from the dataset in \cite{SC_CNN}. We compare SP-CNN with the top two methods from Table \ref{tab:results}, i.e.\ SC-CNN \cite{SC_CNN} and SR-CNN\cite{Xie2015}. Fig. \ref{fig:example_result}(b) is the SP-CNN output. The ground-truth annotated image vs. the detected nuclei locations by SP-CNN, SC-CNN, and SR-CNN are shown in Figs. \ref{fig:example_result}(c), \ref{fig:example_result}(d), \ref{fig:example_result}(e), and \ref{fig:example_result}(f), respectively. This figure also provides further insight into the merits of SP-CNN for a test image. Two parts of each image in Fig.\ \ref{fig:example_result} are magnified for convenience. While `green' zoomed area shows the missed detection by SC-CNN and SR-CNN, `yellow' one shows the wrong detection (FP) of nuclei by those methods. Thanks to shape priors, SP-CNN does not detect them as nuclei, since no nuclei boundaries are detected at those locations.

\section{Conclusion}
We present a deep network for cell nuclei detection in microscopic cellular images. This method uses Shape Priors with Convolutional Neural Networks (SP-CNN) to perform the detection. SP-CNN excels in challenging cases which occur because of limitations in cellular image quality and diversity of nuclear morphology, i.e. varying nuclei shapes, sizes, and overlaps between multiple cell nuclei. This is accomplished by incorporating domain knowledge as informative prior information into the network. SP-CNN demonstrates improvement in cell nuclei detection task over the state-of-the-art and most recent deep learning based nuclei detection methods.

\begin{spacing}{0.9}
\bibliographystyle{IEEEbib}
\bibliography{strings,refs}
\end{spacing}

\end{document}